\pdfoutput=1

\documentclass[11pt]{article}

\usepackage[]{acl}

\usepackage{times}
\usepackage{latexsym}
\usepackage{booktabs}
\usepackage{multirow}
\usepackage{multicol}
\usepackage{amsmath}
\usepackage{graphicx}
\usepackage{physics}
\usepackage{pgfplots}
\usetikzlibrary{calc}
\pgfplotsset{compat=1.18} 
\usepackage{makecell}
\usepackage{CJKutf8}
\usepackage{outlines}
\usepackage{hyperref}
\usepackage{subcaption}

\usepackage[T1]{fontenc}

\usepackage[utf8]{inputenc}

\usepackage{microtype}

\title{Multilingual Language Models are not Multicultural: A Case Study in Emotion
}

\author{{\bf Shreya Havaldar, Sunny Rai, Bhumika Singhal, \ Langchen Liu} \\ { \bf Sharath Chandra Guntuku, \& Lyle Ungar} \\
  University of Pennsylvania \\
  \small\texttt{\{shreyah,sunnyrai,bhsingha,langchen,sharathg,ungar\}@upenn.edu} \\}

\begin{document}
\maketitle

\begin{abstract}
Emotions are experienced and expressed differently across the world. In order to use Large Language Models (LMs) for multilingual tasks that require emotional sensitivity, LMs must reflect this cultural variation in emotion. In this study, we investigate whether the widely-used multilingual LMs in 2023 reflect differences in emotional expressions across cultures and languages. We find that embeddings obtained from LMs (e.g., XLM-RoBERTa) are Anglocentric, and generative LMs (e.g., ChatGPT) reflect Western norms, even when responding to prompts in other languages. Our results show that multilingual LMs do not successfully learn the culturally appropriate nuances of emotion and we highlight possible research directions towards correcting this.

\end{abstract}

\section{Introduction}

The global reach of Large Language Models (LMs) today prompts an important question -- \textit{Are multilingual LMs also multicultural?} We are specifically interested in the multicultural behavior of LMs from the lens of emotion. LMs are used for many multilingual tasks that require emotional sensitivity and therefore must be able to reflect cultural variation in emotion. For instance, LM-powered Therapy Bots must delicately adapt the way they speak to patients in different languages \cite{gpt-therapybot}, LMs as creative writing assistants must produce content that will elicit the appropriate emotional response in an author's desired audience \cite{gpt-storytelling}, LMs used for workplace communication must understand the subtleties of interpersonal interaction \cite{gpt-email}, etc.

\begin{figure}[t]
    \centering
    \includegraphics[width=\linewidth]{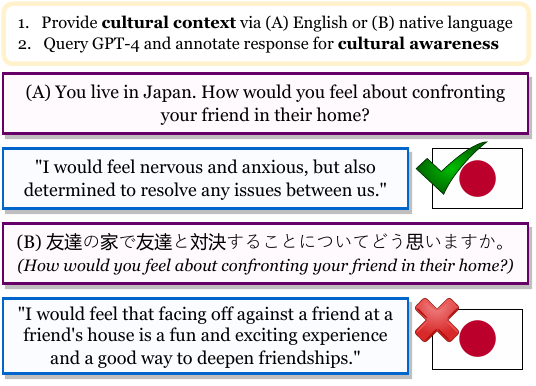}
    \caption{Do LMs always generate culturally-aware emotional language? We prompt GPT-4 to answer "How would you feel about confronting your friend in their home?" like someone from Japan. We provide cultural context either via English (stating "You live in Japan" in the prompt) or via a Japanese prompt. GPT-4 returns two drastically different completions, with the Japanese completion annotated as not culturally appropriate.}
    \label{fig:code-switch}
\end{figure} 

We define cultural variation in emotion as \textit{the nuances in meaning and usage of emotion words across cultures.} For example, in English, we have many different words that express Anger. One can  say "I feel angry," but may also choose to say "frustrated", "irritated", or "furious." The Anger invoked by a baby crying on an airplane is different from the Anger invoked by an unfair grade on an exam; different situations that cause Anger will invoke different language to best express it. These nuances in meaning and usage patterns of emotion words exist differently across cultures \cite{mesquita1997culture,wierzbicka1999emotions}. 

Therefore, there is not a perfect one-to-one mapping between languages for emotion words coupled with their meaning and usage patterns. The direct translation for "I feel frustrated" from English to Chinese (simplified), for example, is \begin{CJK*}{UTF8}{gbsn}"我感到沮丧"\end{CJK*}. However, in a situation where a native English speaker would likely say "I feel frustrated," a native Chinese speaker may use a different phrase than \begin{CJK*}{UTF8}{gbsn}"我感到沮丧"\end{CJK*}, based on situation, context, and the cultural norms of emotion expression in China. 

As we rely on multilingual LMs today for emotionally sensitive tasks, they must reflect this cultural variation in emotion. However, the widely-used multilingual LMs are trained on Anglocentric corpora and encourage alignment of other languages with English \cite{reimers-2020-multilingual-sentence-bert}, both implicitly and explicitly, during training. The key problem in this approach to building multilingual LMs is that any form of alignment destroys a model's ability to encode subtle differences, like the difference between “I feel frustrated” in the United States and \begin{CJK*}{UTF8}{gbsn}"我感到沮丧"\end{CJK*} in China.

In this paper, we investigate whether widely-used multilingual LMs reflect cultural variation in emotion.  We select four high-resource written languages, two Western and two Eastern, to focus on in this work -- English, Spanish, Chinese (Simplified), and Japanese. 

Specifically, we investigate two facets of LMs: embeddings and language generation.

\begin{outline}[enumerate]
\vspace{-0.1cm}
    \1  Emotion embeddings  
    \vspace{-0.1cm}
        \2 \textbf{Does implicit and explicit alignment in LMs inappropriately anchor emotion embeddings to English?} We compare embeddings from monolingual, multilingual, and aligned RoBERTa models.
        \2 \textbf{Do emotion embeddings reflect known psychological cultural differences?} We project embeddings onto the Valence-Arousal plane to visualize American vs. Japanese differences in Pride and Shame.
    \vspace{-0.6cm}
    \1  Emotional language generation 
    \vspace{-0.1cm}
        \2 \textbf{Do LMs reflect known psychological cultural differences?} We analyze whether GPT-3 probabilities encode American vs. Japanese differences in Pride and Shame.
        \2 \textbf{Do LMs provide culturally-aware emotional responses?} We prompt GPT-3.5 and GPT-4 with scenarios that should elicit varied emotional responses across cultures and conduct a user study to assess response quality.
\end{outline}

We make our code public~\footnote{\url{https://github.com/shreyahavaldar/Multicultural_Emotion/}} and encourage researchers to utilize the analyses outlined in this work as a baseline to measure the cultural awareness of future multilingual models.

\section{Related Work}

A large body of work in NLP focuses on detecting emotion in multilingual text . However, a major oversight in this line of research is that \textit{it treats emotion as culturally invariant}. Work from \citet{xlm-emo-wassa} gathers a corpus of annotated social media data from 19 languages, but uses machine translation to transfer annotations from one language to another, assuming that translation correctly captures emotional variation. Work from \citet{sven-lexica} generates lexica to analyze emotion across 91 languages, relying on translations from English lexica and assuming that the affective state of parallel words will be identical.

 Psychologists  have characterized emotion as having multiple components -- an emotional experience, a physiological response, and a behavioral response tendency \cite{kensinger2006processing}.  Each of these components vary from culture to culture \cite{mesquita1997culture}, a complexity completely ignored when emotion is treated as a \textit{static, transferable label} on an utterance of text. Using machine translation to transfer emotion labels between languages incorrectly assumes that emotion is experienced identically across cultures.

Others have also observed that LMs can fail to account for cultural context and variation. \citet{gpt-culturally-aware} find that ChatGPT strongly aligns with American values. \citet{wvs-embeddings} use word embeddings to globally measure human values across cultures, and find that these values overlap more when measured via data in English vs. native languages. \citet{probing-values} probe multilingual LMs and discover weak alignment with the cultural values reflected by these LMs and established values surveys.

In this paper, we focus on emotion, showing a wider variety of Anglocentric anchoring by elucidating the underlying mechanisms of this alignment. We investigate emotion embeddings and LM probabilities, as well as affective language generated from multilingual LMs.

\section{Investigating Emotion Embeddings}

Many tasks in multilingual NLP utilize embeddings from pre-trained LMs such as XLM-RoBERTa~\cite{roberta-xlm} and mBERT~\cite{devlin2018bert}. Researchers fine-tune these models for downstream tasks, relying on their learned representations of words and concepts.

We scope our investigation to embeddings from the widely used XLM-RoBERTa models. XLM-RoBERTa was trained on text that includes parallel and comparable corpora (e.g., Wikipedia) in multiple languages. The nature of Wikipedia, which has topic-aligned articles in different languages, causes \textit{implicit alignment} in training. Worse, XLM-RoBERTa variants trained via multilingual knowledge distillation \cite{reimers-2020-multilingual-sentence-bert} enforce English sentences and their translations to map to the same point in embedding space, giving \textit{explicit alignment} of other languages with English.

This section investigates the effect of alignment -- both implicit and explicit -- by analyzing emotion embeddings from monolingual, multilingual, and aligned RoBERTa models (See Table \ref{tab:models-experiment-one}).
We further investigate whether this anchoring impacts our ability to visualize known cultural differences (e.g. differences between Pride and Shame in the US vs. Japan \cite{tsai2006cultural}) when projecting embeddings into the two-dimensional Valence-Arousal plane \cite{russell1980circumplex}. 

\subsection{Does implicit and explicit alignment inappropriately anchor emotion embeddings to English?}
\label{sec:embedding_alignment}

We analyze whether implicitly aligned embeddings become Anglocentric by comparing emotion embeddings from XLM-RoBERTa to emotion embeddings learned in a parallel, monolingual setting. We further analyze explicit alignment by comparing embeddings from vanilla XLM-RoBERTa to an explicitly aligned variant of XLM-RoBERTa \cite{reimers-2020-multilingual-sentence-bert}.

\begin{figure}[t]
    \centering
    \includegraphics[width=0.9\linewidth]{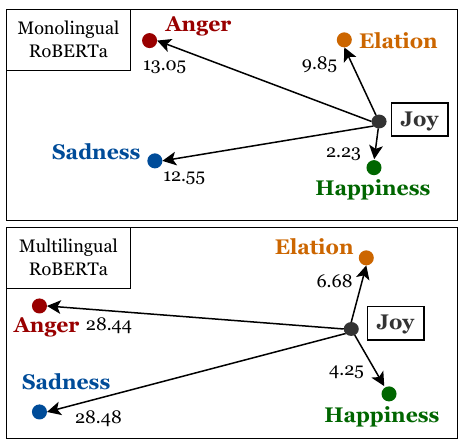}
    \caption{We determine the similarity between the embeddings of monolingual Joy and multilingual Joy by comparing the distances from Joy to other emotions embeddings in both settings. Specifically, we calculate the correlation between $<13.05, 9.85, 12.55. 2.23>$ and $<28.44, 6.68, 28.48, 4.25>$ to infer similarity.}
    \label{fig:nearest1}
\end{figure}

\paragraph{Distance-Based Similarity}
How do we compare the emotion embeddings of two models? Let us take Joy, one of the six Ekman emotions~\cite{ekman1999basic}, as an example -- can we compare the similarity of embeddings from two models for the phrase "I feel joy"? \footnote{We prepend each emotion word with the phrases "I feel" and "I am" to add context and circumvent polysemy when generating embeddings for analysis.} A direct numerical comparison is challenging, as we would need to align the embedding spaces of these two models and possibly distort the Joy embeddings. Taking this into account, we pose the following solution:

The more similar two models are, the more similarly we expect them to embed the same phrases in embedding space. For example, let us embed phrases x, y, and, z using Model A and Model B. This gives us the embedding vectors $\Vec{x}_A, \Vec{y}_A, \Vec{z}_A$ and $\Vec{x}_B, \Vec{y}_B, \Vec{z}_B$ respectively. Figure~\ref{fig:nearest1} illustrates this, showing the embeddings of Joy, Anger, Elation, Sadness, and Happiness using a monolingual and multilingual RoBERTa model. 

If Model A and Model B have embedded phrases x, y, and z in a similar way, then we expect to see a high correlation between the numerical distances $x \rightarrow y, x \rightarrow z,$ and $y \rightarrow z$ in the respective embedding spaces of Model A and B. We calculate the correlation between the following two vectors:

$<\norm{\Vec{x}_A-\Vec{y}_A},\norm{\Vec{x}_A-\Vec{z}_A},\norm{\Vec{y}_A-\Vec{z}_A}>$ 

$<\norm{\Vec{x}_B-\Vec{y}_B},\norm{\Vec{x}_B-\Vec{z}_B},\norm{\Vec{y}_B-\Vec{z}_B}>$

\noindent to inform how similar the embeddings of x, y, and, z are between Model A and Model B. 

Using this idea, we can compare the \textit{distances} from "I feel joy" to other contextualized emotion phrases (e.g. "I feel anger", "I feel happiness", etc.) in embedding space A to those same distances in embedding space B. For example, if the monolingual and multilingual RoBERTa models shown in Figure~\ref{fig:nearest1} have learned similar representations of Joy, then we can expect to see a high Pearson correlation between the vectors $<13.05, 9.85, 12.55. 2.23>$ and $<28.44, 6.68, 28.48, 4.25>$. We use this distance-based similarity metric to answer the following three questions:

\begin{enumerate}
\vspace{-0.1cm}
    \itemsep=0em
    \item Do implicitly aligned multilingual LMs embed emotion words differently than monolingual LMs?
    \item Do implicitly aligned multilingual LMs embed emotion words in an Anglocentric way?
    \item Does explicit alignment further anchor multilingual emotion embeddings to English?
\end{enumerate}

\begin{table*}[htb]
\centering
\small
\begin{tabular*}{\linewidth}{@{\extracolsep{\fill}}lrrrr}
\toprule
& \multicolumn{1}{c}{Mono vs. Multi} & \multicolumn{2}{c}{English vs. Non-English} & \multicolumn{1}{c}{Aligned vs. Unaligned} \\ \midrule
Language (L) &  $\Bar{{r}}(L_{mono},L_{multi})$ & $\Bar{r}(En, L)_{mono}$ & $\Bar{r}(En, L)_{multi}$ &  $\Bar{r}(L_{algn}, L_{unalgn})_{multi}$ %
\\ \midrule
English (En) & \textbf{0.758} (0.35) & --- & --- &\textbf{ 0.483} (0.22) \\
Spanish & 0.318$^*$ (0.20) & 0.222$^*$ (0.14) & \textbf{0.628}$^*$ (0.36) & 0.280$^*$ (0.19) \\ %
Chinese & 0.378$^*$ (0.10) & 0.213$^*$ (0.12) & \textbf{0.437}$^*$ (0.35) & 0.102$^*$ (0.06) \\ %
Japanese & 0.332$^*$ (0.18) & 0.055$^*$ (0.09) & \textbf{0.485}$^*$ (0.39) & 0.332$^*$ (0.18)\\ %
\bottomrule
\end{tabular*}
\caption{We report the average distance-based similarity across 271 emotions for each of our experiments (standard deviation given in parentheses). $^*$indicates the difference in mean correlation between English vs. non-English settings (for Mono vs. Multi, Aligned vs. Unaligned) and monolingual vs. multilingual settings (for English vs. Non-English) is statistically significant ($p<0.05$); we compute this using an independent t-test. See Table \ref{tab:models-experiment-one} for models used in each setting.
}
\label{tab:distance-based-similarity}
\end{table*}

\paragraph{Do implicitly aligned multilingual LMs embed emotion words differently than monolingual LMs?} We compare the emotion representations from \textit{monolingual} and \textit{multilingual} RoBERTa models across English, Spanish, Chinese, and Japanese. We select the four monolingual RoBERTa models most downloaded on Huggingface, additionally ensuring the four models selected have the same number of parameters. Table~\ref{tab:models-experiment-one} contains additional details on the models used in our experiments.\footnote{We note that differences in training data for the monolingual RoBERTa models affect how these models are able to capture emotion. However, it is important to investigate LMs actively used in NLP research rather than explicitly creating a perfectly parallel set of monolingual models.} 

Figure~\ref{fig:nearest1} illustrates this experiment. In practice, we use a list of 271 emotions \cite{emotion-list} for our distance-based similarity computation. Additionally, to account for variance in descriptions of experiencing emotion, we average the embedding of two contextualized phrases for each emotion -- "I feel \textit{<emotion>}" and "I am \textit{<emotion>}".   

For non-English languages, we machine translate the two contextualized English phrases for each emotion (e.g. a representation of Joy in English is the average of the embeddings of "I feel joy" and "I am joyful". The representation of Joy in Spanish is the average of the embeddings "siento alegría" and "soy alegre", etc.). In order to ensure quality, we have native speakers evaluate a subset of the machine-translated emotion phrases, and we find that translation does yield sufficient results. 

We then apply our distance-based similarity metric to compare the monolingual and multilingual emotion embeddings across languages. The "Mono vs. Multi" column in Table \ref{tab:distance-based-similarity} shows the average distance-based similarity across all 271 emotions. The lower similarities for non-English languages indicate that \textit{XLM-RoBERTa embeds non-English emotions differently compared to monolingual models}. We can thus say that multilingual LMs do not preserve the embedding space of monolingual non-English LMs.

\begin{figure}[t]
    \centering
    \includegraphics[width=0.9\linewidth]{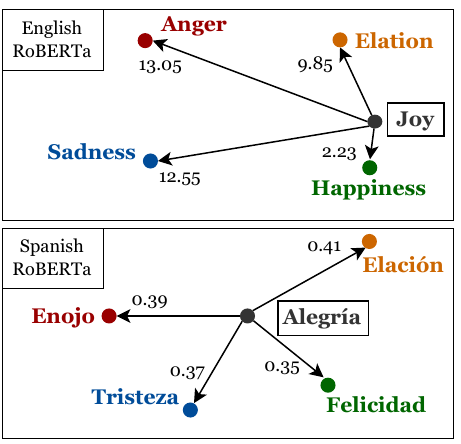}
    \caption{We compare the similarity between the embeddings of Joy in English and Joy(Alegría) in Spanish by comparing the distances from Joy to other emotion embeddings in both languages. Specifically, we calculate the correlation between $<13.05, 9.85, 12.55. 2.23>$ and $<0.39, 0.41, 0.37, 0.35>$ to infer similarity.}
    \label{fig:nearest2}
\end{figure}

\paragraph{Do implicitly aligned multilingual LMs embed emotion words in an Anglocentric way?} We compare the emotion representations of \textit{English} vs. \textit{non-English} languages. We apply our distance-based similarity metric to measure the similarity between English and non-English emotion representations in two settings --  monolingual and multilingual. Figure~\ref{fig:nearest2} illustrates this experiment.

The "English vs. Non-English" columns in Table \ref{tab:distance-based-similarity} show the average distance-based similarity between English and non-English emotion embeddings across all 271 emotions, in monolingual and multilingual settings respectively. Results reveal low similarity between non-English and English emotion embeddings in monolingual space. \textit{In a multilingual setting, however, the non-English emotion embeddings become more similar to English ones.} This suggests that implicit alignment in multilingual LMs anchors non-English emotion embeddings to their English counterparts.  

\paragraph{Does explicit alignment further anchor multilingual emotion embeddings to English?} We compare emotion embeddings from an \textit{unaligned} RoBERTa model to a RoBERTa model trained via \textit{forced alignment} across English, Spanish, Chinese, and Japanese \cite{reimers-2020-multilingual-sentence-bert}.

The average distance-based similarity between aligned and unaligned emotion embeddings across all 271 emotions is shown in column "Aligned vs. Unaligned" in Table~\ref{tab:distance-based-similarity}. \textit{Emotion embeddings from explicitly aligned models are most similar to unaligned embeddings in English}, indicating explicitly aligned embedding space fails to preserve the structure of non-English embedding spaces.

\noindent \paragraph{Finding 1:} Multilingual LMs embed non-English emotion words differently from their monolingual counterparts, whereas English emotion embeddings are more stable and similar in all settings. We demonstrate that \textit{implicit and explicit alignment in multilingual LMs anchor non-English emotion embeddings to English emotions.} All observed trends persist under ablation studies on the effect of distance metric and correlation function (see Appendix~\ref{appendix:exp1}).

\subsection{Do emotion embeddings reflect known psychological cultural differences?}
\label{exp_2}

 Though emotion embeddings from multilingual LMs are Anglocentric, we nonetheless investigate whether they encode any information about known cultural variation in emotion. Prior work \cite{tsai2017ideal, russell1989cross} underlines the differences in emotional expression across cultures, and often illustrates these differences via the circumplex model of affect \cite{russell1980circumplex}. The circumplex model assumes all emotions can be classified along two independent dimensions -- \textit{arousal} (the magnitude of intensity or activation) and \textit{valence} (how negative or positive).

Pride and Shame are two widely researched emotions when investigating cultural differences in emotional expression. \cite{lewis2010cultural,wong2007cultural}. Shame is expressed more commonly and has a desirable affect in Eastern cultures compared to Western cultures. Similarly, Pride is openly expressed in Western cultures whereas Eastern cultures tend to inhibit the feeling of Pride \cite{lim2016arousal}. Moreover, these proclivities are deeply ingrained in society and thus acquired at a very young age \cite{furukawa2012cultural}.
 
 For our experiments, we consider the US and Japan, as the subtle differences in expression of Pride and Shame between these two cultures are well-studied \cite{kitayama2000culture, tsai2006cultural}. We project emotion embeddings from English and Japanese onto the Valence-Arousal plane to visualize whether multilingual LMs capture the expected differences in Pride and Shame. When comparing the embeddings, we expect to specifically observe:
 \begin{enumerate}
 \vspace{-0.1cm}
    \itemsep=0em
     \item The embedding for English Pride should have a more positive valence. \textit{(as Pride is more accepted in the US than Japan)} \cite{furukawa2012cultural}
     \item The embedding for English Shame should have a more negative valence. \textit{(as Shame is more embraced in Japan than the US)} \cite{furukawa2012cultural}
     \item The embeddings for English Pride should have higher arousal \textit{(as Pride is more internally and culturally regulated in Japan than the US)} \cite{lim2016arousal}
 \end{enumerate}

\begin{figure}[h]
    \centering
    \definecolor{darkgreen}{RGB}{0,100,0}
    \begin{tikzpicture}
        \begin{axis}[
            width=\linewidth,
            height=160pt,
            title={Ekman Emotions on the V-A Plane},
            title style = {align=center},
            xlabel={\textbf{valence}},
            xlabel style={at={(0.5, -0.1)}, anchor=north},
            ylabel={\textbf{arousal}},
            ylabel style={at={(-0.15, 0.5)}, anchor=north},
            xmin=-1.5, xmax=1.5,
            ymin=-1.2, ymax=1.5,
            scatter/classes={
                Fear={mark=*, draw=magenta, fill=magenta},
                Anger={mark=*, draw=purple, fill=purple},
                Joy={mark=*, draw=violet, fill=violet},
                Sadness={mark=*, draw=cyan, fill=cyan},
                Disgust={mark=*, draw=darkgreen, fill=darkgreen},
                Surprise={mark=*, draw=orange, fill=orange},
                Axis={mark=*, draw=lightgray, fill=lightgray}
            },
            grid=both,
            scatter, only marks,
            scatter src=explicit symbolic,
            visualization depends on={value \thisrow{displaylabel} \as \displaylabel},
            nodes near coords*={\displaylabel},
            every node near coord/.append style={
                font=\small,
                color=black,
                anchor=west,
                inner sep=3pt,
            }
        ]
        \addplot table[meta=label] {
            x y label displaylabel
            -0.8508593 1.045894 {Fear} {Fear}
            -0.67413247 0.76769996 {Anger} {Anger}
            0.9546092 -0.21090859 {Joy} {Joy}
            -1.1178665 -0.27373743 {Sadness} {Sadness}
            -0.69370687 0.60925367 {Disgust} {Disgust}
            0.3570112 1.2101021 {Surprise} {Surprise}
            1 0 {Axis} {\textit{PV}}
            -1 0 {Axis} {\textit{NV}}
            0 1 {Axis} {\textit{HA}}
            0 -1 {Axis} {\textit{LA}}
        };
        \end{axis}
    \end{tikzpicture}
    \vspace{-0.05in}
    \caption{The six Ekman emotions projected onto the Valence-Arousal plane. We replicate the circumplex model of affect, enabling visualization and theoretical analysis of multi-dimensional emotion embeddings.}
    \label{fig:Ekman_VA}
\end{figure}
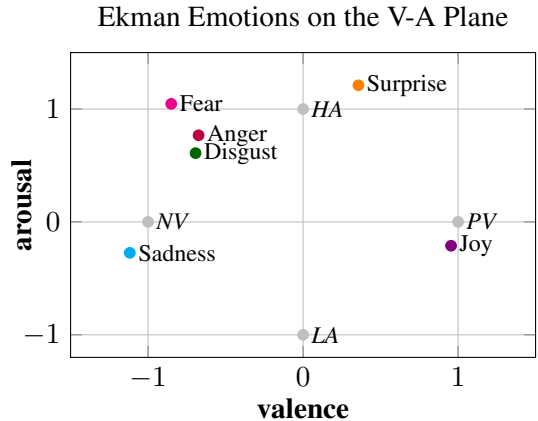 

\paragraph{Projection into the Valence-Arousal plane} In order to define the valence and arousal axes, we first generate four axis-defining points by averaging the contextualized embeddings of the emotions listed in Table~\ref{tab:axis_definitions}. This gives us four vectors in embedding space that best represent positive valence ($PV$)  negative valence ($NV$), high arousal ($HA$), and low arousal ($LA$). We can now project any emotion embedding onto the plane defined by the valence axis ($NV \rightarrow PV$) and the arousal axis ($LA \rightarrow HA$). We give a more formal, mathematical description of this projection method in the Appendix~\ref{appendix:exp2}. Figure~\ref{fig:Ekman_VA} shows the six Ekman emotions \cite{ekman1999basic} projected into the Valence-Arousal plane, indicating that our projection method successfully recreates the circumplex.

\begin{figure}[t]
    \centering
    \begin{tikzpicture}
        \begin{axis}[
            width=\linewidth,
            height=150pt,
            title={Pride \& Shame on the V-A Plane},
            title style = {align=center},
            xlabel={\textbf{valence}},
            xlabel style={at={(0.5, -0.1)}, anchor=north},
            ylabel={\textbf{arousal}},
            ylabel style={at={(-0.15, 0.5)}, anchor=north},
            xmin=-1.5, xmax=1.5,
            ymin=-1.5, ymax=1.25,
            grid=both,
        ]
        \addplot+[only marks, mark=*, mark options={scale=1, draw=lightgray,fill=lightgray}, nodes near coords, point meta=explicit symbolic, nodes near coords style={anchor=south, font=\tiny, color=black}] coordinates {
            (1, 0) [\textit{PV}]
            (-1, 0) [\textit{NV}]
            (0, 1) [\textit{HA}]
            (0, -1) [\textit{LA}]
        };       
        \addplot+[only marks, mark=*, mark options={scale=1, draw=blue,fill=blue}, nodes near coords, point meta=explicit symbolic, nodes near coords style={anchor=north, font=\small, color=black}] coordinates {
            (-0.98046577, -0.08031465) [Shame]
            (0.7600758, 0.047394346) [Pride]
        };      
        \addplot+[only marks, mark=*, mark options={scale=1, draw=red, fill=red}, nodes near coords, point meta=explicit symbolic, nodes near coords style={anchor=east, font=\small}, color=black] coordinates {
            (-0.85903883, -0.08132414) [Shame]
            (0.6943474, 0.047786526) [Pride]
        };        
        \legend{
            {},
            {\small{En (Aligned)}},
            {\small{Ja (Aligned)}}
        }
    \end{axis}
    \end{tikzpicture}
    \begin{tikzpicture}
        \begin{axis}[
            width=\linewidth,
            height=150pt,
            title style = {align=center},
            xlabel={\textbf{valence}},
            xlabel style={at={(0.5, -0.1)}, anchor=north},
            ylabel={\textbf{arousal}},
            ylabel style={at={(-0.15, 0.5)}, anchor=north},
            xmin=-1.5, xmax=1.5,
            ymin=-1.5, ymax=1.5,
            grid=both,
        ]
        \addplot+[only marks, mark=*, mark options={scale=1, draw=lightgray,fill=lightgray}, nodes near coords, point meta=explicit symbolic, nodes near coords style={anchor=north, font=\tiny, color=black}] coordinates {
            (1, 0) [\textit{PV}]
            (-1, 0) [\textit{NV}]
            (0, 1) [\textit{HA}]
            (0, -1) [\textit{LA}]
        };       
        \addplot+[only marks, mark=*, mark options={scale=1, draw=blue,fill=blue}, nodes near coords, point meta=explicit symbolic, nodes near coords style={anchor=west, font=\small, color=black}] coordinates {
            (-1.2192397, 0.182077146) [Shame]
            (0.85887146, -0.22163567) [Pride]
        };      
        \addplot+[only marks, mark=*, mark options={scale=1, draw=red, fill=red}, nodes near coords, point meta=explicit symbolic, nodes near coords style={anchor=west, font=\small}, color=black] coordinates {
            (-1.0286912, -0.29807085) [Shame]
            (0.7637955, 0.17435119) [Pride]
        };        
        \legend{
            {},
            {\small{En (Unaligned)}},
            {\small{Ja (Unaligned)}}
        }
    \end{axis}
    \end{tikzpicture}
    \vspace{-0.1in}
    \caption{We project English and Japanese Pride and Shame embeddings into the Valence-Arousal plane. We use an aligned (top) and unaligned (bottom) RoBERTa model to embed the contextualized emotions. In both cases, we do not see all of our hypotheses confirmed.}
    \label{fig:PS_VA_Aligned}
\end{figure}
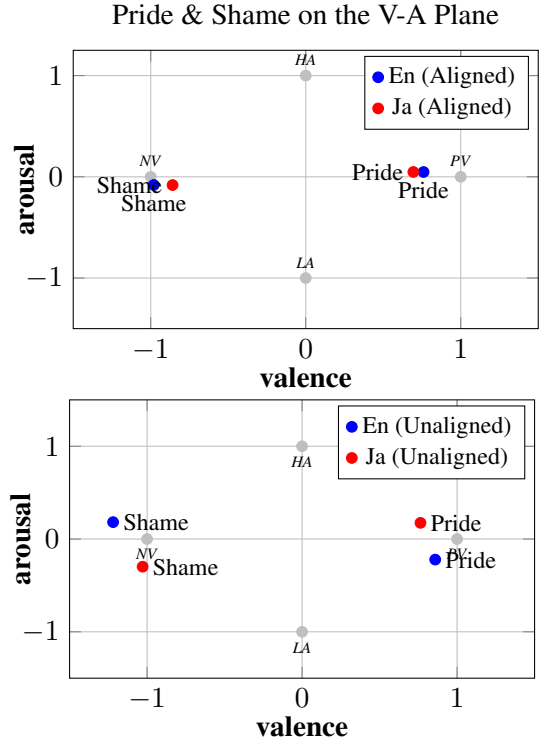

To visualize Pride and Shame in the Valence-Arousal plane, we manually translate the axis-defining emotions to Japanese and average the English and Japanese points of each axis category to define \textit{multilingual valence and arousal axes}. We then project the contextualized sentence embeddings "I am proud" and "I am ashamed" in English and Japanese. We experiment with both aligned and unaligned RoBERTa models; these plots are shown in Figure~\ref{fig:PS_VA_Aligned}. 

Looking at the plots, we observe that English Pride is slightly higher in valence than Japanese Pride, and English Shame is slightly lower in valence than Japanese Shame. This does serve as a weak confirmation of the first two hypotheses. However, we do not observe English Pride to have higher arousal than Japanese Pride. This discrepancy suggests our results are inconclusive, and we cannot confirm whether multilingual RoBERTa encodes cultural variation in English vs. Japanese Pride and Shame.

\noindent \paragraph{Finding 2:} By projecting emotion embeddings into the Valence-Arousal plane, we show that \textit{LMs are not guaranteed to encode the nuances in meaning and usage of emotion words across cultures.} Researchers who utilize embeddings from multilingual LMs for emotion-related tasks assume these pre-trained models have learned adequate representations of emotion across languages. However, implicit and explicit alignment during training causes multilingual LMs to ignore the subtle differences in emotion expression across cultures. 

\begin{figure*}[t]
    \centering
    \begin{tikzpicture}
        \begin{axis}[
            width=\textwidth,
            height=125pt,
            title={"I received an award in front of my coworkers. I feel \_\_\_."},
            title style = {align=center},
            xlabel={\textbf{Emotion}},
            xlabel style={at={(0.5, -0.1)}, anchor=north},
            ylabel={\textbf{GPT-3 Log Probability}},
            ymin=-30, ymax=-12.5,
            xtick={1, 2, 3, 4},
            xticklabels={proud, ashamed, happy, embarrassed},
            xticklabel style={rotate=0, anchor=north, align=center},
            yticklabel style={align=center},
            grid=both,
            minor tick num=1,
            nodes near coords,
            point meta=explicit symbolic,
            legend style={at={(0.97,0.95)}, anchor=north east,legend columns=2,font=\footnotesize},
            every node near coord/.append style={black, anchor=west, inner sep=5pt, font=\footnotesize}
        ]
        \addplot+[only marks, nodes near coords] coordinates {
            (1, -17.834) [-17.83]
            (2, -24.926) [-24.93]
            (3, -23.863) [-23.86]
            (4, -23.890) [-23.89]
        };
        \addplot+[only marks, nodes near coords, mark=*] coordinates {
            (1, -14.236) [-14.24]
            (2, -22.559) [-22.56]
            (3, -20.497) [-20.50]
            (4, -27.832) [-27.83]
        };
        \legend{English,Japanese}
        \end{axis}
    \end{tikzpicture}
    \vspace{-0.2in}
    \caption{A comparison of GPT-3 sentence completion probabilities in English and Japanese. We show the log probabilities for the sentence "I feel X." following the scenario "I received an award in front of my coworkers." and test emotion words associated with Pride or Shame in English and Japanese. Contrary to cultural expectation, we do not observe a pattern where Pride words have a higher likelihood in English or Shame words have a higher likelihood in Japanese.}
    \label{fig:logprob}
\end{figure*}
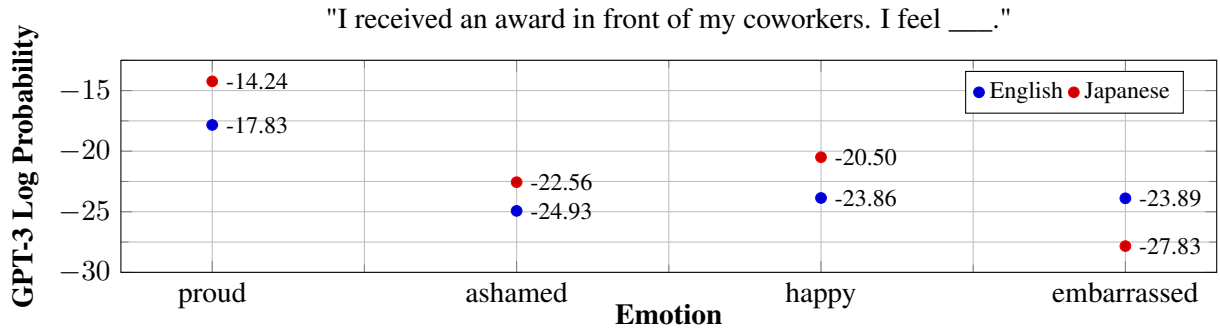

\section{Investigating multilingual LM generation}

We now turn from investigating embeddings to analyzing language generated by Language Models (GPT-3, GPT-3.5, and GPT-4) to see if multilingual LM completions reflect cultural variation in emotion. In order for LMs to be used for tasks that require emotional sensitivity, their responses must align with cultures' socio-cultural norms \cite{genesee1982social}; generated text must reflect users' cultural tendencies and expected affect \cite{tsai2017ideal}.

We first analyze token-level completion probabilities from GPT-3, to see if they reflect cultural differences between American and Japanese Shame and Pride. We then prompt GPT-3.5 and GPT-4 in English and non-English languages to respond to scenarios that should elicit different emotional responses across cultures and assess their cultural appropriateness in a small-scale user study.

\subsection{ Do LMs reflect known psychological cultural differences?}
\label{exp_3}

Continuing our example of English vs. Japanese Pride and Shame, we evaluate whether this known cultural difference is reflected in OpenAI's GPT-3. 

We design a set of 24 prompts (See Table \ref{tab:gpt3_scenarios}) for GPT-3 (\texttt{davinci}) based on six scenarios that would invoke a combination of Pride and Shame in the form \texttt{<context><feeling>}. For example, "I received an award in front of my coworkers. I feel proud." One might feel proud for receiving an award or embarrassed for being publically praised. We then prompt GPT-3 using various \texttt{<context><feeling>} prompts, and analyze the log probability of each token of the prompt. Finally, we sum the log probability of each token in the \texttt{<feeling>} sentence to get a sense of how likely the  \texttt{<feeling>} is to follow the \texttt{<context>}. Based on cultural norms about how one would react in situations that elicit both Pride and Shame, we expect to see a higher probability for "I feel happy" and "I feel proud" in English, and a higher probability for "I feel embarrassed" and "I feel ashamed" in Japanese across scenarios. 

Figure~\ref{fig:logprob} shows the results of this for the prompt "I received an award in front of my coworkers. I feel \_\_\_." where we test two Pride words: "proud", "happy", and two Shame words: "ashamed", and "embarrassed". We replicate this experiment in Japanese, and compare the summed log probabilities of "I feel \_\_\_." between English and Japanese across emotions. The full results, along with the remaining prompts are given in Appendix Table~\ref{tab:gpt3_scenarios}.
Analyzing the results across six scenarios (see Appendix~\ref{appendix:exp3}), we do not see any consistent evidence that Pride is more likely to be expressed in English or Shame is more likely to be expressed in Japanese. In Figure~\ref{fig:logprob}, for example, we see contradicting results for "proud", "happy", and "embarrassed".

\noindent \paragraph{Finding 3:} These results suggest that \textit{GPT-3 lacks knowledge of Pride and Shame and the norms surrounding their expression in the US and Japan.} This is a major limitation; such a failure to capture cultural variation is likely to limit both the utility and applicability of LMs in downstream emotionally-sensitive tasks.

\subsection{Do LMs provide culturally-aware emotional responses?}

\begin{table*}[hbtp]
    \centering
    \small
    \begin{tabular}{p{0.95\textwidth}}
    \toprule
    \textit{Prefix} \\ \midrule
   You are a helpful chatbot. Your goal is to answer my questions like you are a human capable of feelings and emotions. \{\textbf{You live in China.}\} Answer the following question using a single sentence that begins with "I would feel...". \\ \midrule 
   \textit{Questions} \\ \midrule
   Your neighbor offers you food. How would you feel turning it down? \\
   How would you feel if your parents moved in with you right after you were married? \\
   How would you feel if your guests chose to keep their shoes on when entering your home? \\
   \bottomrule
    \end{tabular}
    \caption{Example prompts (\texttt{<prefix>+<question>}) designed to evaluate whether GPT-3.5 and GPT-4 can adapt to account for cultural variation in emotion. In our first set of experiments, we include the bold sentence "You live in China." and prompt GPT in English. In our second set of experiments, we do NOT include the bold sentence, and instead provide cultural context by translating our \texttt{<prefix>+<question>} prompt to Chinese. The full set of questions is given in Appendix Table~\ref{tab:prompts_culturalnorms_emotions}.}
    \label{tab:prompts_completion}
\end{table*}

To further investigate whether LM completions reflect cultural norms, we conduct a small-scale user study to see if GPT-3.5 and GPT-4 are capable of appropriately adapting when prompted in different languages. Annotators assess whether the completions parallel the accepted emotional responses associated with the user's culture.

\paragraph{Prompting with cultural context}
Prior psychological research has detailed scenarios that reveal how emotional expressions vary across cultures \cite{mesquita2022between}. We use this work to design a set of 19 questions (see Table \ref{tab:prompts_culturalnorms_emotions}) that should elicit different emotional responses across cultures. For example, the question "How would you feel if your guests chose to keep their shoes on when entering your home?" would likely elicit a different response from someone culturally American vs. Chinese.

We use these scenarios to prompt GPT-3.5  (\texttt{gpt-3.5-turbo}) and GPT-4 (\texttt{gpt-4}) in the form \texttt{<prefix>+<question>} (see Table~\ref{tab:prompts_completion}). In order to include cultural context and coax the LM into returning a culturally appropriate emotional response, we experiment with providing cultural context in two ways (using Chinese as an example):
\begin{enumerate}
\vspace{-0.1cm}
    \itemsep=0em
    \item \textit{via English} -- we add the sentence "You live in China." to the prefix. The LM returns an English completion.
    \item \textit{via Native Language} -- we translate the \texttt{<prefix>+<question>} prompt to Chinese. The LM returns a Chinese completion.
\end{enumerate}
Figure~\ref{fig:code-switch} details these two cultural context modes and how they may cause conflicting LM responses.

\paragraph{User Study} To assess the quality of the LM completions, we perform a small-scale user study using eight volunteers, consisting of four pairs fluent in English, Spanish, Chinese, and Japanese respectively. We ask our volunteers to annotate GPT-3.5 and GPT-4's responses for cultural awareness along two axes - \textit{linguistic norms} (how you would expect a native speaker to talk), and \textit{cultural norms} (what you would expect a native speaker to say). As these two norms are deeply correlated, annotators are instructed to take both of these dimensions into account and give a single rating to each completion. We use a scale of 1-7, where 7 indicates the LM's response is fully expected of a native speaker.

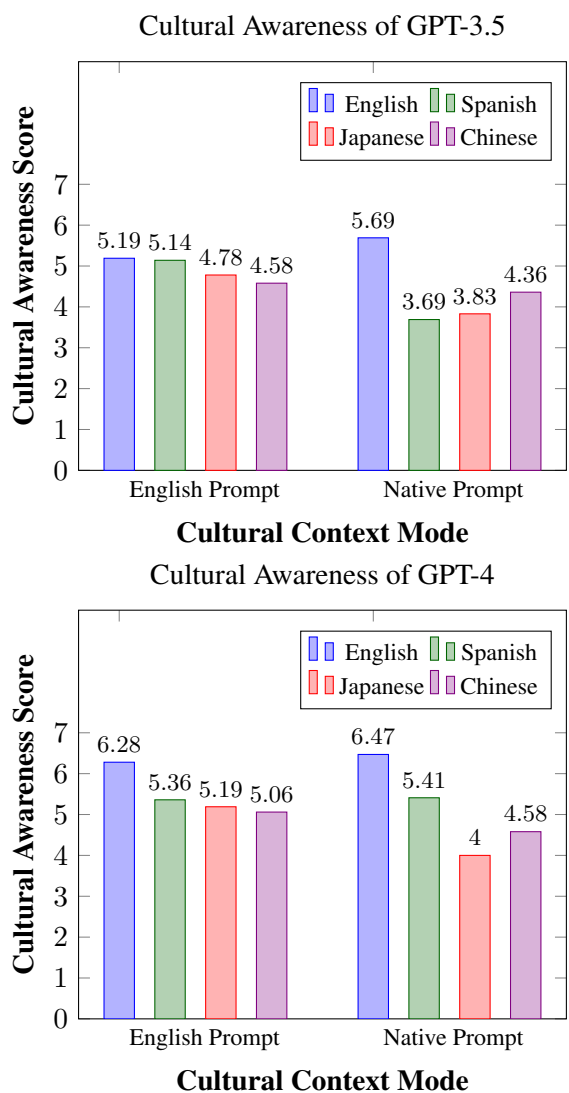
\begin{figure}[hbtp]
    \centering
    \begin{subfigure}[t]{0.5\textwidth}
        \centering
        \begin{tikzpicture}
        \definecolor{darkgreen}{RGB}{0,100,0}
        \definecolor{darkyellow}{RGB}{180,100,0}
        \begin{axis}[
            title={Cultural Awareness of GPT-3.5},
            ybar,
            bar width=0.4cm,
            width=\linewidth,
            height=7cm,
            symbolic x coords={En1, Es1, Ja1, Zh1, Space, En2, Es2, Ja2, Zh2},
            xtick={En1, En2},
            xticklabels={English Prompt, Native Prompt},
            nodes near coords,
            nodes near coords align={vertical},
            nodes near coords style={font=\footnotesize, black},
            xlabel=\textbf{{Cultural Context Mode}},
            ylabel=\textbf{{Cultural Awareness Score}},
            ymin=0,
            ymax=10,
            ytick={0,1,2,3,4,5,6,7},
            legend style={at={(0.97,0.95)}, anchor=north east,legend columns=2,font=\footnotesize},
            enlarge x limits=0.1,
            xtick align=inside,
            x tick label style={rotate=0, anchor=north west, font=\footnotesize},
            /pgf/bar shift=0pt
        ]
        \addplot+[draw=blue, fill=blue!30!white] coordinates {(En1,5.19)};\   
        \addplot+[draw=darkgreen, fill=darkgreen!30!white] coordinates {(Es1,5.14)};
        \addplot+[draw=red, fill=red!30!white] coordinates {(Ja1,4.78)};
        \addplot+[draw=violet, fill=violet!30!white] coordinates {(Zh1,4.58)};
        
        \addplot+[draw=blue, fill=blue!30!white] coordinates {(En2,5.69)};\  
        \addplot+[draw=darkgreen, fill=darkgreen!30!white] coordinates {(Es2,3.69)};
        \addplot+[draw=red, fill=red!30!white] coordinates {(Ja2,3.83)};
        \addplot+[draw=violet, fill=violet!30!white] coordinates {(Zh2,4.36)};

        \legend{English, Spanish, Japanese, Chinese}
        \end{axis}
    \end{tikzpicture}
        \label{Fig:SubLeft}
    \end{subfigure}
    \begin{subfigure}[t]{0.5\textwidth}
        \centering
        \begin{tikzpicture}
        \definecolor{darkgreen}{RGB}{0,100,0}
        \definecolor{darkyellow}{RGB}{180,100,0}
        \begin{axis}[
            title={Cultural Awareness of GPT-4},
            ybar,
            bar width=0.4cm,
            width=\linewidth,
            height=7cm,
            symbolic x coords={En1, Es1, Ja1, Zh1, Space, En2, Es2, Ja2, Zh2},
            xtick={En1, En2},
            xticklabels={English Prompt, Native Prompt},        
            nodes near coords,
            nodes near coords align={vertical},
            nodes near coords style={font=\footnotesize, black},
            xlabel=\textbf{{Cultural Context Mode}},
            ylabel=\textbf{{Cultural Awareness Score}},
            ymin=0,
            ymax=10,
            ytick={0,1,2,3,4,5,6,7},
            legend style={at={(0.97,0.95)}, anchor=north east,legend columns=2,font=\footnotesize},
            enlarge x limits=0.1,
            xtick align=inside,
            x tick label style={rotate=0, anchor=north west, font=\footnotesize},
            /pgf/bar shift=0pt
        ]
        \addplot+[draw=blue, fill=blue!30!white] coordinates {(En1,6.28)};\   
        \addplot+[draw=darkgreen, fill=darkgreen!30!white] coordinates {(Es1,5.36)};
        \addplot+[draw=red, fill=red!30!white] coordinates {(Ja1,5.19)};
        \addplot+[draw=violet, fill=violet!30!white] coordinates {(Zh1,5.06)};
        
        \addplot+[draw=blue, fill=blue!30!white] coordinates {(En2,6.47)};\  
        \addplot+[draw=darkgreen, fill=darkgreen!30!white] coordinates {(Es2,5.41)};
        \addplot+[draw=red, fill=red!30!white] coordinates {(Ja2,4.00)};
        \addplot+[draw=violet, fill=violet!30!white] coordinates {(Zh2,4.58)};

        \legend{English, Spanish, Japanese, Chinese}
        \end{axis}
    \end{tikzpicture}
        \label{Fig:SubRight}
    \end{subfigure}
    \caption{Average cultural awareness scores across annotations for GPT-3.5 and GPT-4 completions in each language. We observe a consistently higher quality of English completions, and poor performance of Eastern languages compared to Western, especially when prompted using the \textit{Native Language} context mode.}
    \label{fig:gpt_annotation}
\end{figure}

Across languages, we observe a high agreement within each pair of volunteers. Figure~\ref{fig:gpt_annotation} details the average score across annotators and questions for GPT-4 and GPT-3.5 completions. We provide the annotator agreement statistics in Appendix Table~\ref{tab:annotator_agreements}. Analyzing the completions and annotations, we notice some interesting trends:
\begin{itemize}
\vspace{-0.1cm}
    \itemsep=0em
    \item We see a large difference in quality between the LM responses returned using the two cultural context prompting modes (even though the questions are identical.) 
    \item For Chinese and Japanese, the LM returns a less culturally-appropriate response using the \textit{Native Language} cultural context mode.
    \item English completions are the most culturally-aware across languages, and English response quality is unaffected by cultural context mode.
\end{itemize}

\paragraph{Finding 4: } GPT-3.5 and GPT-4 fail to infer that a prompt in a non-English language suggests a response that aligns with the linguistic and cultural norms of a native speaker. Additionally, the LM completions reflect culturally appropriate emotion much better in Western languages than Eastern.

\section{Conclusion}
We find that multilingual models fail to fully capture cultural variations associated with emotion, and predominantly reflect the cultural values of the Western world.  Emotion embeddings from multilingual LMs are anchored to English, and the text completions generated in response to non-English prompts are not in tune with the emotional tendencies of users' expected culture. For instance, when GPT-4 is prompted in Japanese, it responds as an American fluent in Japanese but unaware of Japanese culture or values.

Our results caution against blindly relying on emotion representations learned by LMs for downstream applications. Using machine translation to transfer labels or utilizing multilingual LMs in a zero-shot setting for unseen languages has risks -- the multilingual representations of emotion learned by these models do not perfectly reflect how their corresponding cultures express emotion. 

\paragraph{Future Research Directions} Our paper motivates the need for future work that transcends current Anglocentric LMs. This could take the form of higher performing, non-English models in a monolingual setting, or of multilingual models trained on more linguistically and culturally balanced corpora. Future work should additionally investigate whether state-of-the-art monolingual models in non-English languages succeed in encoding the respective culture's norms. Furthermore, we encourage the evaluation of multilingual models on benchmarks that measure cultural awareness in addition to standard metrics.

\section{Limitations}
We only analyze four high-resource languages in this study, our analysis could have benefited from more languages, especially low-resource ones. Additionally, we only analyze Japanese and English Pride/Shame as a known cultural difference; analyzing other differences could provide stronger results. We perform a small user study, and our work could have benefited from a larger-scale study with more annotators and completions analyzed. 

We recognize the added complexity of investigating Pride embeddings from a culture where explicit expressions of Pride are discouraged; we note this may be a contributing factor to our results indicating that LMs do not reflect the culturally appropriate nuances of Shame and Pride. Additionally, we acknowledge that the experiments outlined in this paper are specific to investigating cultural awareness from the lens of emotion. These experiments are not easily applicable to measuring cultural awareness from different perspectives; therefore, results may not be generalizable.

At a higher level, we equate \textit{language} with \textit{culture}. Psychologists have observed higher cultural similarities within languages than between them \cite{stulz2003culture}, however, we recognize there are variations within the populations that speak each language. For example, Spanish is spoken by people in Spain, Mexico, and other countries, each having a unique and varied culture.

\section{Ethical Considerations}
Although culturally-aware multilingual LMs are critical in uses such as therapy, storytelling, and interpersonal communication, these are possible misuses for nefarious purposes - persuasion, misinformation generation, etc. Additionally, our analyses behave as if China, Japan, Spain, and the United States are a single culture with a single set of cultural norms. In reality, this is not the case; we recognize there are huge variations in the way people view emotion within each of these cultures. 

\bibliography{custom}
\bibliographystyle{acl_natbib}

\appendix
\setcounter{table}{0}
\renewcommand{\thetable}{A\arabic{table}}
\setcounter{figure}{0}
\renewcommand{\thefigure}{A\arabic{figure}}

\section{Distance-based Similarity Experiments: Additional Details}
\label{appendix:exp1}

 \begin{table}[b]
     \centering
     \small
     \begin{tabular}{ll}
        \toprule
         \textbf{Axis Anchor} & \textbf{Russell Emotions} \\ \midrule
         Positive valence (PV) &  \makecell[l]{Happy, Pleased, Delighted, \\ Excited, Satisfied}  \\ 
         Negative valence (NV) &  \makecell[l]{Miserable, Frustrated, Sad, \\ Depressed, Afraid}  \\
         High arousal (HA) &  \makecell[l]{Astonished, Alarmed, Angry, \\ Afraid, Excited}  \\
         Low arousal (LA) &  \makecell[l]{Tired, Sleepy, Calm, \\ Satisfied, Depressed}  \\
         \bottomrule
     \end{tabular}
     \caption{Emotions used to define the valence and arousal axis anchors for projection into the Valence-Arousal plane. We select the 5 emotions from the circumplex closest to each axis point.}
     \label{tab:axis_definitions}
 \end{table}

\begin{table*}[!t]
    \centering
    \small
    \begin{tabular}{lllp{14.5em}}
    \toprule
    \textbf{Language \& Setting} & \textbf{Model Name} & \textbf{Downloads} & \textbf{Training Data}  \\ \midrule
    Monolingual English & \makecell[l]{\texttt{roberta-base}\\ \cite{roberta-base}} & 7.77M &  BookCorpus, Wikipedia, Common Crawl(News), OpenWebText, Stories  \\
    Monolingual Spanish & \makecell[l]{\texttt{bertin-roberta-base-spanish}\\ \cite{spanish-roberta}} & 2.67k & Common Crawl \\
    Monolingual Chinese & \makecell[l]{\texttt{chinese-roberta-wwm-ext}\\ \cite{chinese-roberta}} & 113k & Wikipedia, Encyclopedia, News, Web QA data \\
    Monolingual Japanese & \makecell[l]{\texttt{japanese-roberta-base}\\ \cite{rinna_pretrained2021}} & 36.2k & Common Crawl, Wikipedia \\
    \midrule
    Multilingual, Unaligned & \makecell[l]{\texttt{xlm-roberta-base}\\ \cite{roberta-xlm}} & 18.4M & Common Crawl, Wikipedia \\ 
    Multilingual, Aligned & \makecell[l]{\texttt{paraphrase-multilingual-} \\\texttt{mpnet-base-v2}\\ \cite{reimers-2019-sentence-bert}} & 293k & Common Crawl, Wikipedia, Aligned Paraphrasing Corpus \\
    \bottomrule
    \end{tabular}
    \caption{RoBERTa models used in our experiments for each setting: monolingual, multilingual, and aligned. For each model, we provide the number of monthly downloads by Huggingface users (as of April 2023) and a high-level description of the data used for training. All models have 125M parameters.}
    \label{tab:models-experiment-one}
\end{table*}

Table~\ref{tab:models-experiment-one} gives details on the RoBERTa models we use in each setting -- monolingual, multilingual, and aligned -- for all experiments in this paper.

We find no clear pattern in certain emotions being more or less problematic across languages. Our machine translations of 271 English emotions give 247, 210, and 246 unique emotions for Spanish, Chinese, and Japanese respectively. 

In order to test the robustness of the experiments outlined in section \ref{sec:embedding_alignment}, we experiment with other distance and correlation metrics in our distance-based similarity calculations. Table~\ref{tab:distance-based-similarity-ablation} shows results for our distance-based similarity experiments where we replace Euclidean distance with cosine similarity, and results where we replace Pearson correlation with Spearman's rank.

\begin{table*}[!t]
\centering
\small
\begin{tabular*}{\linewidth}{@{\extracolsep{\fill}}lrrrr}
\toprule
& \multicolumn{1}{c}{Mono vs. Multi} & \multicolumn{2}{c}{English vs. Non-English} & \multicolumn{1}{c}{Aligned vs. Unaligned} \\ \midrule
Language (L) &  $\Bar{{r}}(L_{mono},L_{multi})$ & $\Bar{r}(En, L)_{mono}$ & $\Bar{r}(En, L)_{multi}$ &  $\Bar{r}(L_{algn}, L_{unalgn})_{multi}$ %
\\ \midrule
\textit{Using cosine distance} &&&& \\ \midrule
English (En) & \textbf{0.752} & --- & --- &\textbf{ 0.468} \\
Spanish & 0.290$^*$ & -0.219$^*$ & \textbf{0.647}$^*$ & 0.252$^*$ \\ %
Chinese & 0.338$^*$ & -0.223$^*$ & \textbf{0.454}$^*$ & 0.067$^*$ \\ %
Japanese & 0.303$^*$ & -0.05$^*$ & \textbf{0.490}$^*$ & 0.287$^*$\\ \midrule %
\textit{Using Spearman's rank} \\ \midrule
English (En) & \textbf{0.652} & --- & --- &\textbf{ 0.488} \\
Spanish & 0.339$^*$ & 0.248$^*$ & \textbf{0.567}$^*$ & 0.307$^*$ \\ %
Chinese & 0.377$^*$ & 0.223$^*$ & \textbf{0.418}$^*$ & 0.162$^*$ \\ %
Japanese & 0.334$^*$ & 0.059$^*$ & \textbf{0.460}$^*$ & 0.353$^*$\\ %
\bottomrule
\end{tabular*}
\caption{We report the average distance-based similarity across 271 emotions for each of our experiments, using cosine distance and Spearman's rank correlation. $^*$indicates the difference in mean correlation between English vs. non-English settings (for Mono vs. Multi, Aligned vs. Unaligned) and monolingual vs. multilingual settings (for English vs. Non-English) is statistically significant ($p<0.05$); we compute this using an independent t-test. See Table \ref{tab:models-experiment-one} for models used in each setting. We see that our observed trends persist despite ablation.
}
\label{tab:distance-based-similarity-ablation}
\end{table*}

\begin{table*}
\centering
    \small
    \begin{tabular*}{\linewidth}{@{\extracolsep{\fill}}lllr}
    \toprule
    \textbf{Language} & \textbf{GPT Model} & \textbf{Cultural Context Mode} & \textbf{Agreement} \\ \midrule
    & \texttt{gpt-3.5-turbo} & \textit{English} & 0.785 \\
    English & \texttt{gpt-3.5-turbo} & \textit{Native Language} & 0.705 \\
    & \texttt{gpt-4} & \textit{English} &  0.823\\
    & \texttt{gpt-4} & \textit{Native Language} & 0.673 \\ \midrule
    & \texttt{gpt-3.5-turbo} & \textit{English} & 0.547 \\
    Spanish & \texttt{gpt-3.5-turbo} & \textit{Native Language} & 0.662 \\
    & \texttt{gpt-4} & \textit{English} & 0.559 \\
    & \texttt{gpt-4} & \textit{Native Language} & 0.776 \\ \midrule
    & \texttt{gpt-3.5-turbo} & \textit{English} & 0.665 \\
    Chinese & \texttt{gpt-3.5-turbo} & \textit{Native Language} & 0.609 \\
    & \texttt{gpt-4} & \textit{English} & 0.708 \\
    & \texttt{gpt-4} & \textit{Native Language} & 0.749 \\ \midrule
    & \texttt{gpt-3.5-turbo} & \textit{English} &  0.847 \\
    Japanese & \texttt{gpt-3.5-turbo}& \textit{Native Language} & 0.878 \\
    & \texttt{gpt-4} & \textit{English} &  0.843 \\
    & \texttt{gpt-4} & \textit{Native Language} & 0.900 \\
    \bottomrule
    \end{tabular*}
    \caption{GPT models used and annotator agreement (Pearson correlation between each annotator pair) for our user study. We observe high agreement between each annotator pair across languages}
    \label{tab:annotator_agreements}
\end{table*}

\section{Projection into the Valence-Arousal plane}
\label{appendix:exp2}
 
In order to define the valence and arousal axes, we first generate four axis-defining points by averaging the contextualized embeddings ("I feel [emotion]") of the emotions listed in Table~\ref{tab:axis_definitions}. This gives us four vectors in embedding space -- positive valence ($\Vec{v}_{pos}$), negative valence($\Vec{v}_{neg}$), high arousal($\Vec{a}_{high}$), and low arousal($\Vec{a}_{low}$). We mathematically describe our projection function below:
\begin{enumerate}
\vspace{-0.1cm}
    \itemsep=0em
    \item We define the valence axis, $V$, as $\Vec{v}_{pos} - \Vec{v}_{neg}$ and the arousal axis, $A$, as $\Vec{a}_{high} - \Vec{a}_{low}$. We then normalize $V$ and $A$ and calculate the origin as the midpoints of these axes: $(\Vec{v}_{middle}, \Vec{a}_{middle})$. 
    \item We then scale the axes so $\Vec{v}_{pos}$, $\Vec{v}_{neg}$, $\Vec{a}_{high}$, and $\Vec{a}_{low}$ anchor to $(1,0)$, $(-1,0)$, $(0,1)$, and $(0,-1)$ respectively.
    \item We Compute the angle $\theta$ between the valence-arousal axes by solving $\cos\theta =  \frac{{V}\cdot{A}}{\norm{V}\cdot\norm{A}}$
    \item For each embedding vector $\Vec{x}$ in the set $\{x_i\}_{i=1}^n$ we want to project into our defined plane, we compute the valence and arousal components for $x_i$ as follows: \\
    $x_i^{v}= (x_i - \Vec{v}_{middle})\cdot \Vec{V}$\\
    $x_i^{a}= (x_i - \Vec{a}_{middle})\cdot \Vec{A}$.
    \item We calculate the x and y coordinates to plot, enforcing orthogonality between the axes: \\
    $\Tilde{x_i^{v}} = x_i^{v} - x_i^{a}\cdot \cos\theta$\\
    $\Tilde{x_i^{a}} = x_i^{a} - x_i^{v}\cdot \cos\theta$\\
    Finally, we plot $(\Tilde{x_i^{v}}, \Tilde{x_i^{v}})$ in the Valence-Arousal plane.   
\end{enumerate}

In order to define multilingual valence and arousal axes and plot English vs. Japanese Pride and Shame embeddings, we calculate $\Vec{v}_{pos}, \Vec{v}_{neg}, \Vec{a}_{high}, \text{and } \Vec{a}_{low}$ separately for English and Japanese. We then average the axis-defining points between English and Japanese (i.e. $\Vec{v}_{pos} = AVG(\Vec{v}_{pos_en}, \Vec{v}_{pos_ja})$, etc.) so we can project embeddings from two languages into the same plane.

\section{GPT-3 Pride \& Shame Experiments: Additional Details}
\label{appendix:exp3}

\begin{table*}
    \centering
    \small
    \begin{tabular*}{\linewidth}{@{\extracolsep{\fill}}p{7.5cm}rrrr}
    \toprule
    \textbf{Scenario + Language} & \textit{proud} & \textit{happy} & \textit{ashamed} & \textit{embarrassed} \\ \midrule
    \textit{My teacher complimented me in front of the class. I feel X. }\\
    English & \textbf{-22.386} & \textbf{-26.923} & \textbf{-29.947} & \textbf{-25.872} \\
    Japanese & -23.012 & -30.642 & -32.584 & -41.096 \\ \midrule
    \textit{My teacher complimented my friend in front of the class. I expect he feels X.} \\
    English & -36.620\textbf{} & -39.324 & -46.868 & -42.395 \\
    Japanese & \textbf{-25.175} & \textbf{-28.946} & \textbf{-33.690} & \textbf{-33.736} \\ \midrule
    \textit{I received an award in front of my coworkers. I feel X.} \\
    English & -17.834 & -23.863 & -24.926 & \textbf{-23.890} \\
    Japanese & \textbf{-14.236} & \textbf{-20.497} & \textbf{-22.559} & -27.832 \\ \midrule
    \textit{My colleague received an award in front of our coworkers. I expect he feels X.} \\
    English & -29.238 & -33.736 & -39.236 & -34.866 \\
    Japanese & \textbf{-14.017} & \textbf{-18.251} & \textbf{-23.928} & \textbf{-21.340} \\ \midrule
    \textit{My parents praised me in front of my siblings. I feel X. } \\
    English & \textbf{-17.925} & \textbf{-21.063} & \textbf{-24.843} & \textbf{-22.504} \\
    Japanese & -23.258 & -30.589 & -30.610 & -40.588 \\ \midrule
    \textit{My parents praised my brother in front of our siblings. I expect he feels X.} \\
    English & -34.805 & -36.931 & -42.669 & -38.356 \\
    Japanese & \textbf{-30.318} & \textbf{-33.213} & \textbf{-35.10}\textbf{8} &\textbf{ -35.152} \\    
    \bottomrule
    \end{tabular*}
    \caption{All scenarios and results for our GPT-3 Pride \& Shame prompting experiment. Prompts are in the form (\texttt{<context><feeling>}) and designed to describe scenarios that should elicit a combination of Shame and Pride. We evaluate whether GPT-3 encodes cultural differences in Japanese vs. US Pride and Shame by comparing the summed log probabilities of the \texttt{<feeling>} sentence.}
    \label{tab:gpt3_scenarios}
\end{table*}

We provide the full list of scenarios used in Table~\ref{tab:gpt3_scenarios}. We also include the results of our experiment across scenarios. 

We find no empirical evidence of a consistent trend that "I feel ashamed" and "I feel embarrassed" are more likely to be said in Japanese or that "I feel proud" and "I feel happy" are more likely to be said in English. Rather, we observe a trend that the higher log probability for an emotion (between English vs. Japanese) is more dependent on the scenario rather than culture.

\begin{table*}
    \centering
    \small
    \begin{tabular}{p{0.95\textwidth}}
    \toprule
    \textit{Prefixes} \\ \midrule
   \textbf{Prefix 1: } You are a helpful chatbot. Your goal is to answer my questions like you are a human capable of feelings and emotions. You live in China. Answer the following question using a single sentence that begins with "I would feel...". \\ 
   \textbf{Prefix 2: } \begin{CJK*}{UTF8}{gbsn}你是一个有用的聊天机器人。你的目标是回答我的问题，就像你是一个有感情和情感能力的人一样。使用以"我会觉得……"开头的单个句子回答以下问题。\end{CJK*}\textit{(You are a helpful chatbot. Your goal is to answer my questions like you are a human capable of feelings and emotions. Answer the following question using a single sentence that begins with "I would feel...".)}\\ \midrule 
   \textit{All Questions} \\ \midrule
   How would you feel about turning down the food offered by your sibling? \\
    How would you feel about turning down the food offered by your neighbor?\\
    How would you feel when providing critical feedback to your colleague?\\
    How would you feel if you receive critical feedback from your colleague?\\
    How would you feel about confronting a colleague in the office?\\
    How would you feel about confronting your friend in their home?\\
    How would you feel about confronting your friend in your home?\\
    How would you feel about prioritizing your friends over family?\\
    How would you feel about prioritizing your family over friends?\\
    How would you feel if your parents moved in with you right after you were married?\\
    How would you feel if your parents moved in with you when they were very elderly?\\
    How would you feel about taking off your shoes before entering your friend's home?\\
    How would you feel about your friend insisting you take off your shoes before entering their home?\\
    How would you feel if your guests chose to keep their shoes on when entering your home?\\
    How would you feel when interacting with the boss of your supervisor?\\
    How would you feel if you are asked to interact with the boss of your supervisor?\\
    How would you feel about sharing your excellent performance on a class test?\\
    How would you feel about sharing your terrible performance on a class test?\\
   \bottomrule
    \end{tabular}
    \caption{All questions included in our user study. Prompts are in the form (\texttt{<prefix>+<question>}) and designed to evaluate whether GPT-3.5 and GPT-4 can adapt to account for cultural variation in emotion.}
    \label{tab:prompts_culturalnorms_emotions}
\end{table*}

\end{document}